\documentclass[preprint,12pt]{elsarticle}

\usepackage{amssymb}
\usepackage{hyperref}
\usepackage{amsmath}
\usepackage{xcolor}
\usepackage{graphicx}    % for \includegraphics
\usepackage{subcaption}  % for subfigures
\usepackage{caption}     % for \captionsetup
\usepackage{multirow}
\usepackage{pgfgantt}
\usepackage{booktabs} % for better looking tables
\usepackage{adjustbox}
\usepackage[table]{xcolor}
\usepackage{appendix}
\usepackage{epsfig}
\journal{Journal of Pathology Informatics}

\begin{document}

\begin{frontmatter}



\title{Decoding Future Risk: Deep Learning Analysis of Tubular Adenoma Whole-Slide Images} %% Article title

\author[label1]{Ahmed Rahu, MD}
\author[label2]{Brian Shula} 
\author[label3]{Brandon Combs}
\author[label4]{Aqsa Sultana}
\author[label1]{Surendra P. Singh, MD}
\author[label4]{Vijayan K. Asari, PhD}
\author[label3]{Derrick Forchetti, MD}
\affiliation[label1]{organization={Dept. of Pathology},
            addressline={University of Toledo Medical Center},
            city={Toledo},
            postcode={43614},
            state={OH},
            country={USA}}

\affiliation[label2]{organization={Honeywell International Inc.},
            city={South Bend},
            postcode={46628},
            state={IN},
            country={USA}}

\affiliation[label3]{organization={South Bend Medical Foundation},
            city={South Bend},
            postcode={46635},
            state={IN},
            country={USA}}

\affiliation[label4]{organization={Dept. of Electrical and Computer Engineering},
            addressline={University of Dayton},
            city={Dayton},
            postcode={45469},
            state={OH},
            country={USA}}


% \author{} %% Author name

% \affiliation{organization={},%Department and Organization

%% Abstract
\begin{abstract}
\textbf{\textit{Background:}}  Colorectal cancer (CRC) has one of the biggest initiatives for prophylactic screening via colonoscopy and removal of precancerous lesions. Despite this, a subset of patients still develop colorectal cancer (CRC). Currently, microscopic evaluation and post-polypectomy strategies are standardized and may miss low-risk appearing adenomas that harbor a higher risk of neoplasia. Advancements in digital pathology and deep learning offer opportunities to detect subtle histologic features that could improve risk stratification at the pre-cancerous stage.

\noindent \textbf{\textit{Methods:} }This is a retrospective cohort study of patients who underwent screening colonoscopies between 2013 and 2022 with biopsy-confirmed tubular adenomas with low-grade dysplasia. Patients with a known predisposition to CRC were excluded from this study. The remaining patients (with low risk of progression to CRC) were split into two cohorts: a non-progressor group (no CRC development) and a progressor group (subsequent CRC development). The glass slides from H\&E-stained sections were scanned and processed into tiles for analysis. An EfficientNetV2S-based convolutional neural network was trained to classify adenomatous tiles and predict progression risk. Model performance was evaluated at both the tile and whole-slide levels using accuracy, precision, recall, F1-score, and AUROC. Model interpretability was assessed using Gradient-weighted Class Activation Mapping (Grad-CAM).

\noindent \textbf{\textit{Results:}} A total of 335,763 high-quality tiles were analyzed, including 40,514 held-out test tiles. Additionally, 20 whole-slide images (WSIs) not used during training were reserved for independent validation. At the tile level, the model achieved an accuracy of 0.9788, precision of 0.9762, recall of 0.9815, F1-score of 0.9789, and excellent discrimination with an AUROC approaching unity. At the whole-slide level, all 20 held-out WSIs (10 progressors and 10 non-progressors) were correctly classified. Grad-CAM analysis revealed distinct patterns between the two cohorts: progressor-associated tiles highlighted regions of increased architectural complexity, nuclear crowding, elongation, and pseudostratification, while non-progressor tiles emphasized preserved glandular architecture and uniform nuclear spacing.

\noindent \textbf{\textit{Conclusion:}} Deep learning–based analysis of digitized histologic images can identify subtle morphologic features within tubular adenomas with low-grade dysplasia that are associated with future colorectal cancer development. These findings refute the null hypothesis that no machine-detectable predictive features exist in low-grade adenomas and support the potential role of artificial intelligence in objective pre-cancer risk stratification. Such approaches may ultimately contribute to more personalized post-polypectomy surveillance strategies and improved CRC prevention.
\end{abstract}


% \begin{highlights}
% \end{highlights}

%% Keywords

\end{frontmatter}



\section{Introduction}
Advancements in digital pathology and machine learning have revolutionized the way diagnostic and prognostic assessments are conducted in cancer by enabling more objective and reproducible methods in the assessment of tissue morphology. Colorectal cancer (CRC) is a leading cause of cancer-related morbidity and mortality worldwide; it is largely preventable through screening colonoscopy and removal of precancerous lesions [\cite{Kumar2018Robbins}, \cite{Rosai_2011}, \cite{colerectalstats}]. Despite widespread screening initiatives, some patients will develop CRC following the diagnosis and removal of low-grade adenomatous polyps, highlighting a gap in current strategies for risk stratification. 

Tubular adenomas (TAs) are commonly encountered during routine screening and typically are regarded as lesions of limited malignant potential [\cite{Kumar2018Robbins}, \cite{Rosai_2011}]. In view of this, post-polypectomy surveillance recommendations are often standardized rather than individualized [\cite{UPreventive1}]. Generally, accurate prediction of disease progression is essential for effective prophylactic cancer strategies, including screening, surveillance, and early intervention. In the discipline of surgical pathology, distinguishing which low-grade adenomas harbor changes associated with an increased risk of progression to neoplasia shows an unmet clinical need. Accurately predicting the risk of malignant transformation would identify patients who require closer monitoring, enabling more informative, personalized surveillance intervals following screening colonoscopy. 

Traditional histopathologic evaluation relies on visual assessment of architectural and cytological features, including nuclear pseudostratification, nuclear pleomorphism, mitotic figures, and tumor-infiltrating immune cells. However, this process can be subjective and prone to interobserver variability and discordance in guidelines across the globe [\cite{Elsheikh_Kirkpatrick_Fischer_Herbert_Renshaw_2010}, \cite{Vennalaganti_Kanakadandi_Goldblum_Mathur_Patil_Offerhaus_Meijer_Vieth_Odze_Shreyas_etal._2017}, \cite{Elmore_Longton_Carney_Geller_Onega_Tosteson_Nelson_Pepe_Allison_Schnitt_etal._2015}]. This may, in part, suggest that subtle changes in morphological patterns associated with future neoplastic progression may be overlooked. Recent developments in the field of deep learning have shown promise in detecting prognostic and predictive signals that encompass the multiple domains of digital pathology.  

Convolutional neural networks (CNNs) have emerged as a promising tool in digital pathology, enabling the automatic extraction and learning of meaningful features directly from digitized histological images [\cite{korbar2017deeplearningclassificationcolorectalpolyps}, \cite{ultralight}, \cite{Steimetz_Simsek_Saha_Xia_Gupta_2025}]. Prior studies have demonstrated that CNN-based models can achieve high-performance accuracy in tasks such as cancer detection, classification, and grading across a myriad of tissue types [\cite{11102194}, \cite{ Raju_Jayavel_Rajalakshmi_Rajababu_2025}, \cite{Yao_Zhang_Zhou_Liu_2019}, \cite{Kaddes_Ayid_Elshewey_Fouad_2025}]. Unlike conventional image analysis methods that require predefined features, deep learning models can leverage raw image data to identify complex, prognostic patterns that may not be apparent through visual assessment alone \cite{ultralight}, \cite{Steimetz_Simsek_Saha_Xia_Gupta_2025},  \cite{korbar2017deeplearningclassificationcolorectalpolyps}.  

In this study, state-of-the-art deep learning techniques are leveraged to predict malignant transformation directly from digitized tissue samples of tubular adenomas with low-grade dysplasia. Using automated analysis of histopathologic images, this approach evaluates whether machine-detectable histologic features can support objective risk stratification at the pre-cancer stage \cite{MollaHoseyni_Imany_Iranpour_Mehrabani_Seifouri_Rafieipour-Jobaneh_Firuzbakht_Masoudi-Nejad_2025}, \cite{Asiri_Senan_Halawani_Abunadi_Mashraqi_Alshari_2025}. Such an approach has the potential to contribute to improved colorectal cancer (CRC) prevention strategies by refining post-polypectomy surveillance intervals and ultimately enhancing patient care in the digital era  \cite{McCaffrey_Jahangir_Murphy_Burke_Gallagher_Rahman_2024}.  

Based on current histopathological practice \cite{Rosai_2011}, the null hypothesis is that no machine-detectable histological features exist within tubular adenomas with low-grade dysplasia that are predictive of future colorectal cancer development in patients without known predisposing conditions. 

\section{Methodology}
\begin{figure}[h!]
    \centering % Center the entire figure
    % First image
    \begin{subfigure}[b]{0.4\textwidth}
        \captionsetup{labelformat=empty}
    \includegraphics[width=\linewidth]{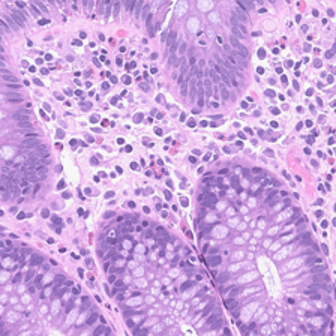}
        \caption{Non-Progressor}
    \end{subfigure}
    \hspace{0.005\textwidth} % optional horizontal spacing
    \begin{subfigure}[b]{0.4\textwidth}
        \captionsetup{labelformat=empty}
    \includegraphics[width=\linewidth]{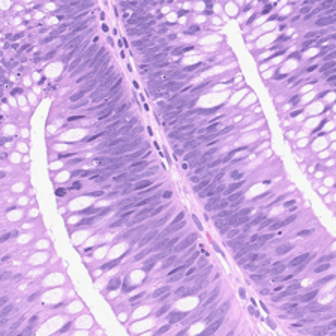}
        \caption{Progressor}
    \end{subfigure}

    \caption{Sample images of the dataset used: non-progressor group (left) and progressor group (right).}
    \label{fig:fig9}
\end{figure}

This retrospective cohort study included patients who underwent screening colonoscopy between 2013 and 2022. Patients were selected based on the presence of biopsy-confirmed tubular adenomas with low-grade dysplasia, and two groups were defined: non-progressors and progressors as shown in Figure \ref{fig:fig9}. Non-progressors were patients who did not develop colorectal cancer during the study period, whereas progressors were patients who developed colorectal cancer during longitudinal follow-up after the index biopsy. Individuals were excluded if they had known risk factors for colorectal cancer, including inflammatory bowel disease (IBD), Lynch syndrome, familial adenomatous polyposis (FAP), or a prior history of colorectal malignancy or high-grade dysplasia. Dysplasia grade was assessed using standardized morphologic criteria, including loss of polarity (specifically loss of basilar orientation), cribriform architecture, intraluminal necrosis, and cytologic atypia.

\begin{table}[h]%% placement specifier
\centering%% For centre alignment of tabular.
\begin{tabular}{l| c r}%% Table column specifiers
%% Tabular cells are separated by &
   &\textbf{ Progressor} & \textbf{Non-Progressor } \\
     &\textbf{Group (n=32)} & \textbf{Group (n=22)} \\%% A tabular row ends with \\
   \hline
 \textbf{ Age (median)} & 79.0 yrs & 69.0 yrs \\
  
 \textbf{ Age (range) }& 54-95 yrs & 54-79 yrs\\
  \textbf{biopsies (mean)} & 1.56 & 2.32 \\
  \textbf{Screening Interval (mean)} & 863 days (2.4 yrs) & 1,659 days (4.5 yrs)
\end{tabular}
%% Use \caption command for table caption and label.
\caption{Demographic and surveillance characteristics of two cases, progressors and non-progressor cohorts }\label{tab1}
\end{table}

Table \ref{tab1} summarizes the demographic and surveillance characteristics of progressor and non-progressor cohorts. Patients underwent one or more screening colonoscopies during the study period, and at least one biopsy specimen from each procedure contained a tubular adenoma with low-grade dysplasia. There were 32 patients in the progressor group (18 Female and 14 Male) and 22 patients in the non-progressors group (11 Female and 11 Male). Patients in the progressor group ranged in age from 54 to 95 years old, with a median age of 79.0 years.
Patients in the non-progressor group ranged in age from 54 to 79 years, with a median age of 69.0 years. The non-progressor group underwent more frequent surveillance, with an average of 2.32 biopsy procedures compared with 1.56 procedures in the progressor group. Consistent with this, the mean screening interval was shorter in the non-progressor group (863 days) than in the progressor group (1,659 days). Notably, all non-progressors had at least two screening events with non-zero intervals, reflecting more consistent longitudinal surveillance, whereas approximately two-thirds of progressors had only a single screening event during the study period. 
Differences in surveillance intensity reflect real-world clinical practice and underscore the need for objective histology-based risk stratification at the time of index biopsy.

% Patients in the non-progressor group ranged in age from 54 to 79 years with a median age of 69.0 years. Patients in the progressor group had an average of 1.56 biopsy procedures during the screening period and in the non-progressor group and an average of 2.32 procedures. The screening interval (time between biopsy procedures) averaged 863 days in the non-progressor group and 1,659 days in the progressor group.

% The non-progressor group is notably younger (median 69 vs 79), had more biopsies on average, and importantly all had at least two screening events with non-zero intervals, suggesting more consistent surveillance compared to the progressor group where two-thirds had only a single screening event. 

\subsection{Histologic Review and Slide Preparation}
All hematoxylin and eosin (H\&E)-stained slides from formalin-fixed, paraffin-embedded (FFPE) tissue sections of tubular adenomas with low-grade dysplasia were prepared by the South Bend Medical Foundation (SBMF). All slides were retrieved from the pathology archives and underwent histopathologic re-review by two board-certified pathologists, one of whom is a subspecialist in gastrointestinal (GI) pathology, to confirm the original diagnoses and exclude high-grade dysplasia and/or carcinoma. Cases with discrepancies in diagnosis or poor tissue preservation were excluded. The selected slides were scanned at 40X magnification using a Leica Biosystems Aperio AT2 Scanner to generate high-resolution whole-slide images (WSIs) in SVS format for computational analysis. 

\subsection{Image Preprocessing and Annotation}
Regions of interest (ROI) corresponding to adenomatous epithelium as well as areas to exclude that were irrelevant to classification were manually annotated by expert pathologists using QuPath \cite{Bankhead_Loughrey_Fernández_Dombrowski_McArt_Dunne_McQuaid_Gray_Murray_Coleman_etal._2017} on a subset of WSIs. Normal mucosa, lymphoid aggregates, and regions with crush artifacts or tangential sectioning were areas labeled for exclusion. These curated annotations were then used to train an EfficientNetV2S-based \cite{Tan_Le_2021} CNN that was pre-trained on ImageNet, to automatically identify adenomatous regions across the remaining dataset.  EfficientNetV2S models were used in this study due to their size, accuracy, and training time compared to other CNN model architectures \cite{Tan_Le_2021}. 

The annotated Whole-slide images (WSIs) were initially reviewed for quality and subjected to preprocessing steps to ensure uniformity of the dataset and optimize input(s) for appropriate model training. WSIs were divided into non-overlapping 1024-pixel tiles to preserve details of the tissue architecture. These were then rescaled to \(224 \times 224\) pixels to meet the input size requirements of the EfficientNetV2S architecture while still maintaining important histological features. Preprocessing included color normalization, image sharpening, and artifact removal to enhance tissue contrast and minimize variability introduced by staining intensity, background noise, and scanning artifacts.  

\subsection{Feature Extraction}
Tile extraction was then performed on the entire set of WSIs, again using the \(1024 \times 1024\) to \(224 \times 224\)-pixel tile resizing. The ROI CNN, trained previously, determined whether to keep a tile or discard it. Following automated ROI generation, all predicted annotations were visually reviewed to ensure accuracy. Tiles not meeting quality standards—due to issues such as tissue folding, edge artifact, or poor scan resolution—were excluded based on inspection of WSI patch location maps. After curation, a total of 335,763 high-quality tiles were retained, consisting of 143,080 from the progressor group and 192,683 from the non-progressor group. These annotated and filtered patches are what formed the ground truth input for CNN training and model validation.  

\section{Experimental Setup and Results}
\subsection{Model Training and Validation }
A second convolutional neural network (CNN) classifier was trained to learn high-dimensional, machine-detectable features from annotated regions of adenomatous epithelium. The network architecture was based on EfficientNetV2 Small (EfficientNetV2S) \cite{Tan_Le_2021}, initialized with ImageNet pre-trained weights and fine-tuned on the curated dataset of whole-slide image (WSI) tiles. A dropout layer (rate = 0.2) was incorporated to mitigate overfitting, and the output layer employed a linear activation with binary cross-entropy loss (from\_logits = True) for improved numerical stability. Model outputs were subsequently transformed using a sigmoid function to obtain class probability estimates.

To enable whole-slide–level evaluation on unseen data, all tiles derived from ten WSIs per class were withheld from training and reserved exclusively for testing. The remaining tiles were class-balanced through random under-sampling of the majority (non-progressor) class and partitioned into training (70\%), validation (15\%), and test (15\%) subsets using stratified sampling to preserve class distributions. This resulted in 189,070 training tiles, 40,514 validation tiles, and 40,514 test tiles.

% A second convolutional neural network (CNN) classifier was trained to extract high-dimensional, machine-detectable features from the aforementioned annotated adenomatous epithelium regions. Once again, the network architecture was based on EfficientNetV2 Small (EfficientNetV2S) \cite{Tan_Le_2021}, initialized with ImageNet pre-trained weights and fine-tuned using the curated dataset of WSI tiles. A Dropout layer with 20\% dropout was added, and the output layer used a linear activation in conjunction with Binary Cross-Entropy Loss with the setting “from\_logits=True” to help with numerical stability. The logits, which are the raw outputs from the model, were passed to a Sigmoid function, which calculates a probability value between zero and one, to indicate the probable Class. 

% The full dataset of 335,763 tiles had all tiles from ten slides from both classes removed, so that model performance testing could be performed at the WSI-level on unseen tiles.  The remaining tiles were balanced by random under-sampling of the majority class (non-progressor group). The resulting dataset was split into training (70\%), validation (15\%), and test (15\%) subsets using stratified sampling to preserve the original class distribution across all partitions. This resulted in 189,070 training tiles, 40,514 validation, and 40,514 test tiles, respectively. 

\subsection{Augmentation Techniques}
\begin{figure}[h!]
\centering
\includegraphics[width=1\textwidth]{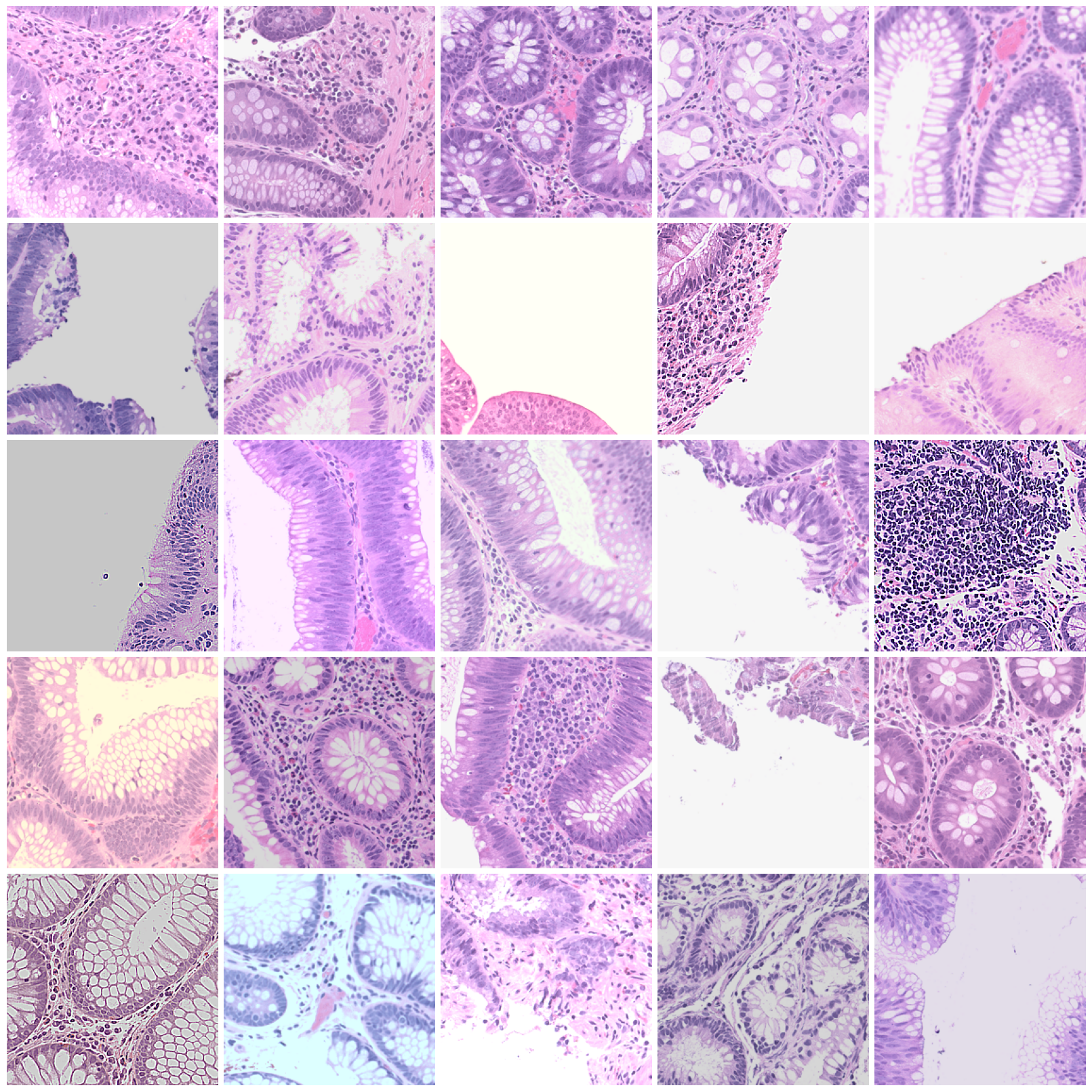}
\caption{Sample Albumentation Tile Augmentations: Representative examples of augmented tiles generated through the Albumentations pipeline. They demonstrate variations in color, contrast, brightness, and sharpness used to improve model robustness and simulate histologic variability across laboratories and slide scanners. }
\label{fig1}
\end{figure}

A data augmentation pipeline was built using the Albumentations \cite{info11020125} library to randomly augment training tiles during each training epoch. These augmentations not only enhance the model’s robustness but also attempt to account for histologic variability. Augmentation techniques included color shifts, color saturations, brightness/contrast enhancements, and image sharpening. Examples of these augmentations are shown in Figure \ref{fig1}. During each training epoch, the pipeline calculates a different augmentation for the same tile, so the model never sees the same variation of a tile twice.

\subsection{Model Training}
Model training was conducted in TensorFlow \cite{TensorFlowDevelopers_2025}, using an EfficientNetV2S backbone initialized with pre-trained ImageNet\cite{imagenet} \cite{NIPS} weights. During the first training phase, all EfficientNet layer weights were frozen, and only the final classification layer weights were updated for 2 epochs with a learning rate of 0.01. In the second phase, the entire network was unfrozen and trained end-to-end for 25 epochs with a reduced initial learning rate of 1e-5, and a decaying learning rate schedule applied after each epoch. A binary cross-entropy loss function was used throughout both training phases. Training took place on an NVIDIA A100 GPU. 
Model optimization was performed using the Adam optimizer, and training was monitored for early stopping based on validation loss to prevent overfitting.
The output of the trained CNN consisted of deep feature embeddings derived from the final convolutional layers, which served as the input for downstream classification tasks and statistical modeling. These embeddings captured complex morphologic patterns in adenomatous tissue that may perhaps be predictive of future colorectal cancer development.

\subsection{Statistical Analysis}
Model performance was evaluated using standard binary classification metrics, including accuracy, precision, recall, F1-score, and area under the receiver operating characteristic curve (AUROC). Classification performance was assessed at both the tile level and the whole-slide level. 

For tile-level evaluation, model outputs were compared against ground truth labels from the test set. For whole-slide inference, predictions for individual tiles were averaged to yield a slide-level probability, with a threshold of 0.5 used to assign class labels. Exploratory model explainability was performed using Gradient-weighted Class Activation Mapping (Grad-CAM) to visualize spatial regions most responsible for each classification. 

\subsection{Experimental Results}
\noindent \textit{Tile--Level Evaluation.} Table \ref{tab2} displays the tile--level confusion matrix of test data. The model predicted 19,773 true negatives (TN), 484 false positives (FP), 374 false negatives (FN), and 19,883 true positives (TP), where a positive prediction refers to the progressor class.
\begin{table}[h!]
\centering

\begin{tabular}{c|cc}
\textbf{Tile-Level Confusion  } & \textbf{Predicted  } & \textbf{Predicted  } \\
\textbf{ Matrix } & \textbf{ Non-Progressor } & \textbf{ Progressor } \\\hline
\textbf{Target Non-Progressor } 
    & \cellcolor{green!35!white}19,773 (TN) 
    & \cellcolor{red!15!white}484 (FP) \\
\textbf{Target Progressor} 
    & \cellcolor{red!20!white}374 (FN)
    & \cellcolor{green!30!white} 19,883 (TP)  \\
\end{tabular}
\caption{Tile--level confusion matrix of test test }\label{tab2}
\end{table}

\begin{figure}[h!]
\centering
\includegraphics[width=0.8\textwidth]{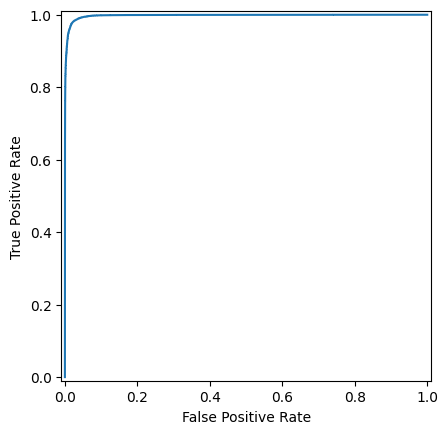}
\caption{ AUROC for the Test Tile Set: Tile-level receiver operating characteristic (ROC) curve generated from the held-out test set (\(n=40,514\) tiles), demonstrating excellent distinction between progressor and non-progressor tiles.   }
\label{fig2}
\end{figure}

The held-out test set consisted of 40,514 unseen tiles, and the model performance is summarized in the Confusion Matrix in Table \ref{tab2}. As shown in Table \ref{tab3}, the model achieved an accuracy of 0.97882, a precision of 0.97624, a recall of 0.98154, and an F-1 score of 0.97888. The AUROC curve is shown in Figure \ref{fig2}, in which the model’s performance is visualized in relation to the ideal classifier curve passing through the point (False Positive Rate = 0.0, True Positive Rate = 1.0).  

\begin{table}[h]%% placement specifier
\centering%% For centre alignment of tabular.
\begin{tabular}{l|cccc}%% Table column specifiers
%% Tabular cells are separated by &
  \textbf{Model} & \textbf{Accuracy} & \textbf{Precision} & \textbf{Recall} & \textbf{F1--Score} \\ %% A tabular row ends with \\
   \hline
\textbf{EfficientNetV2S}& 97.882\% & 0.97624 & 0.98154 & 0.97888
\end{tabular}
%% Use \caption command for table caption and label.
\caption{Performance evaluation of EfficientNetV2S on the dataset }\label{tab3}
\end{table}

\begin{figure}[h!]
\centering

\subfloat[Original WSI]{%
  \includegraphics[height=200pt,width=0.51\linewidth]{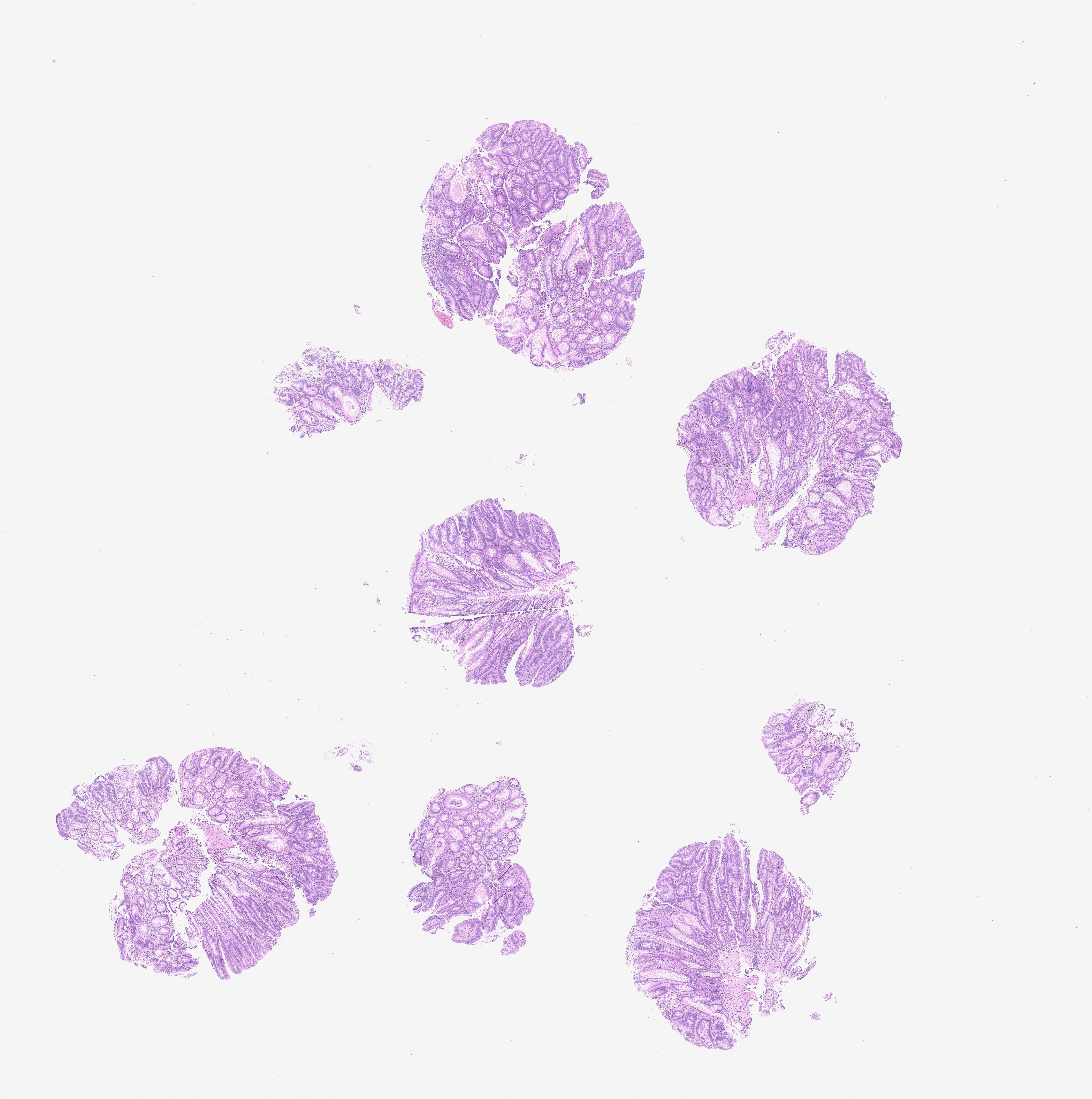}
  \label{fig:3a}
}
\subfloat[Tile Prediction Heatmap ]{%
  \includegraphics[height=200pt,width=0.51\linewidth]{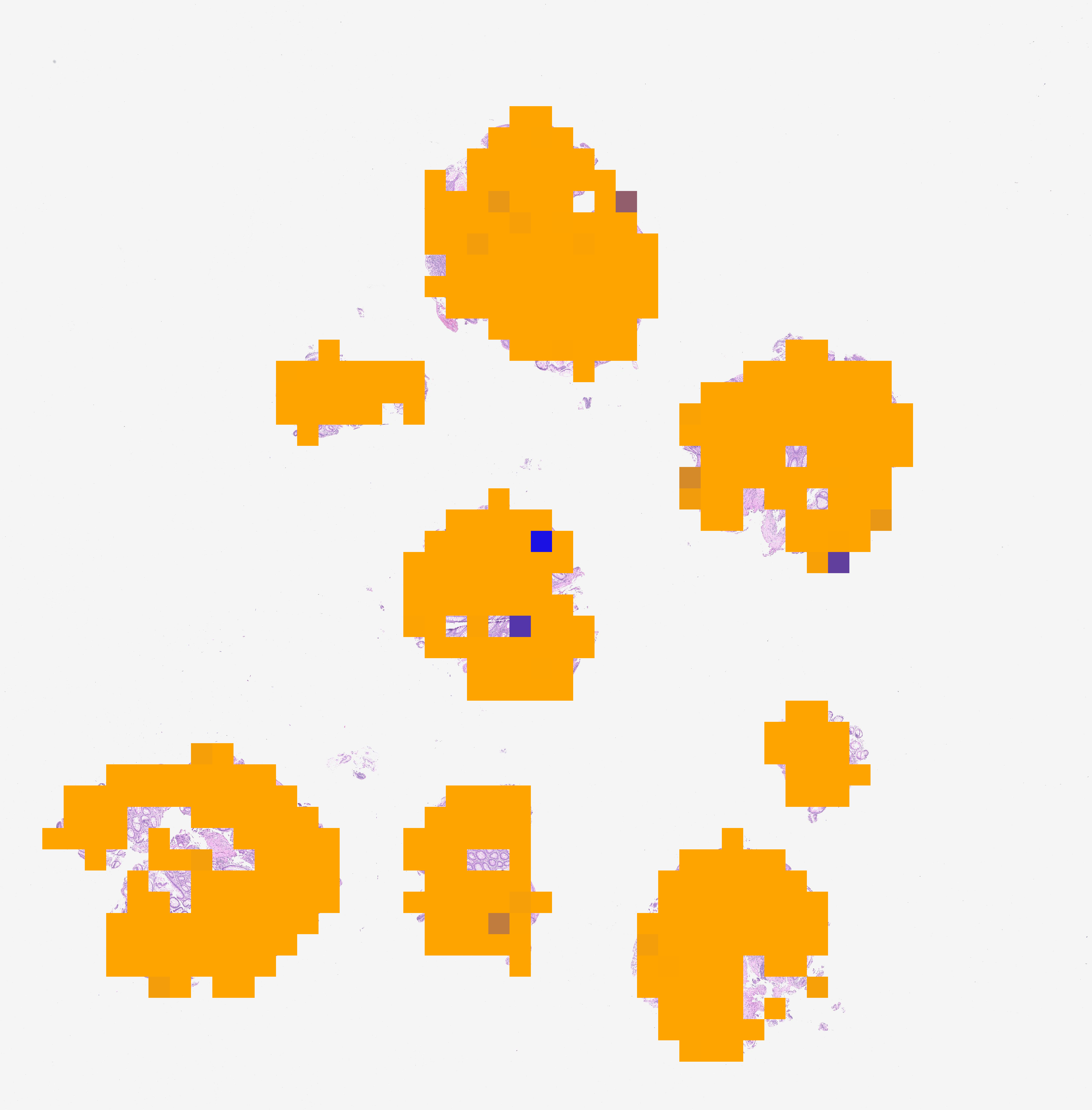}
  \label{fig:3b}
 }

\vspace{0.3cm}

\subfloat[Individual Tiles  ]{%
  \includegraphics[height=200pt,width=0.51\linewidth]{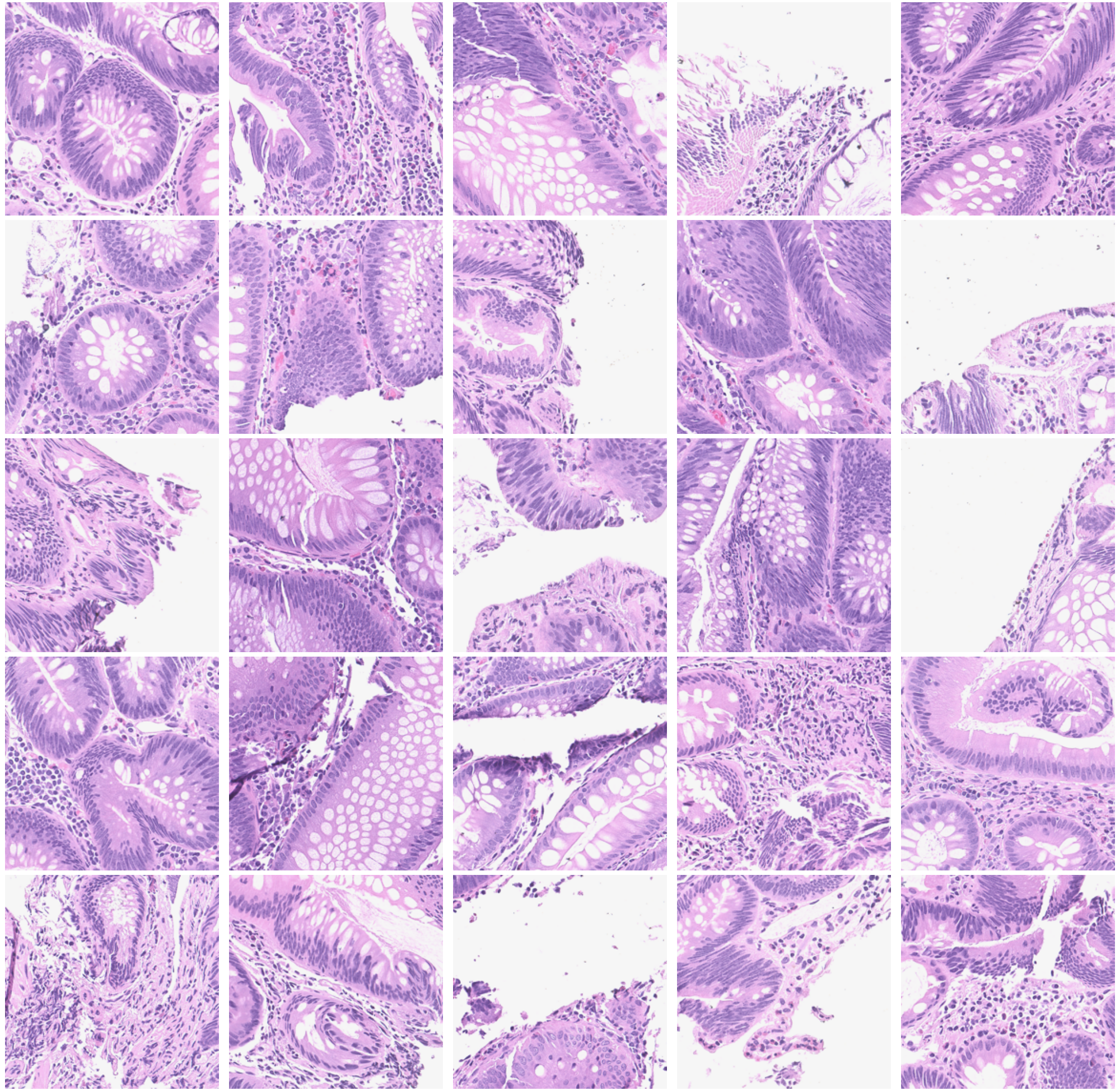}
  \label{fig:3c}
}
\subfloat[Tile Probability Histogram  ]{%
  \includegraphics[height=200pt,width=0.51\linewidth]{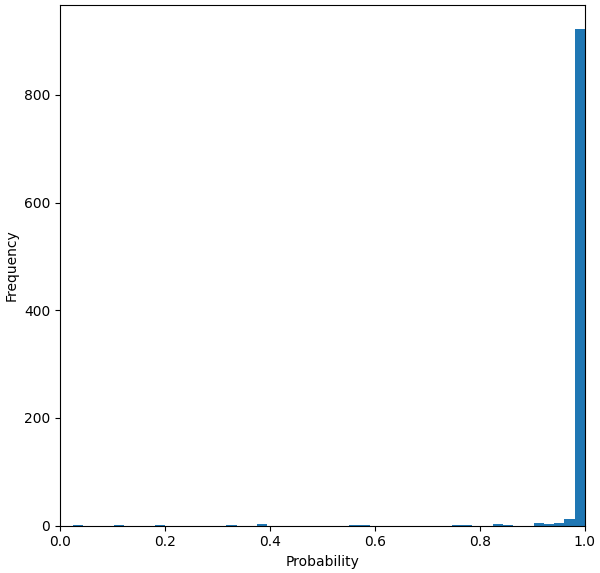}
  \label{fig:3d}
}

\caption{Results for one progressor WSI: (a.) Original WSI (b.) Tile prediction heatmap (c.) Individual tiles (Predicted: Progressor, Target: Progressor) (d.) Histogram of tile probabilities (Target: 1, Predicted: 1)} 
\label{fig_3}
\end{figure}

\noindent \textit{Whole Slide Prediction/Inference.}
To assess model generalization at the whole-slide level, 10 WSIs per class (total n = 20), each from a unique patient, were held out entirely from the training and validation of pipelines. For each WSI, tile-level predictions (probabilities) were generated and averaged to produce a slide-level-class probability. A threshold of 0.5 was applied to determine the class: non-progressor ($p < 0.5$) or progressor ($p \geq 0.5$).  

The model predicted all 20 slides correctly. Figure \ref{fig_3} illustrates a representative example from the progressor group, showing the original WSI (Figure \ref{fig:3a}) alongside a tile-level heatmap(Figure \ref{fig:3b}) of predictions. The heatmap uses an orange-blue continuous contour scale to indicate tile prediction probability in which orange denotes high cancer probability, and blue denotes high non-progressor probability. In this case, the average WSI probability was 0.9915, reflecting high confidence in WSI prediction.  Nearly all tiles are strongly-blue or strongly-orange, indicating confidence in the tile-level predictions as well. Figure \ref{fig:3c} also shows samples of individual tiles from the WSI, as well as a histogram of tile prediction probabilities in Figure \ref{fig:3d}.
\\

\noindent \textit{Grad-CAM: Feature Importance Heatmaps/Model Interpretability.}

\begin{figure}[h!]
\centering

\subfloat[Non-Progressor Group – Grad-CAM Heatmap]{%
  \includegraphics[width=0.65\linewidth]{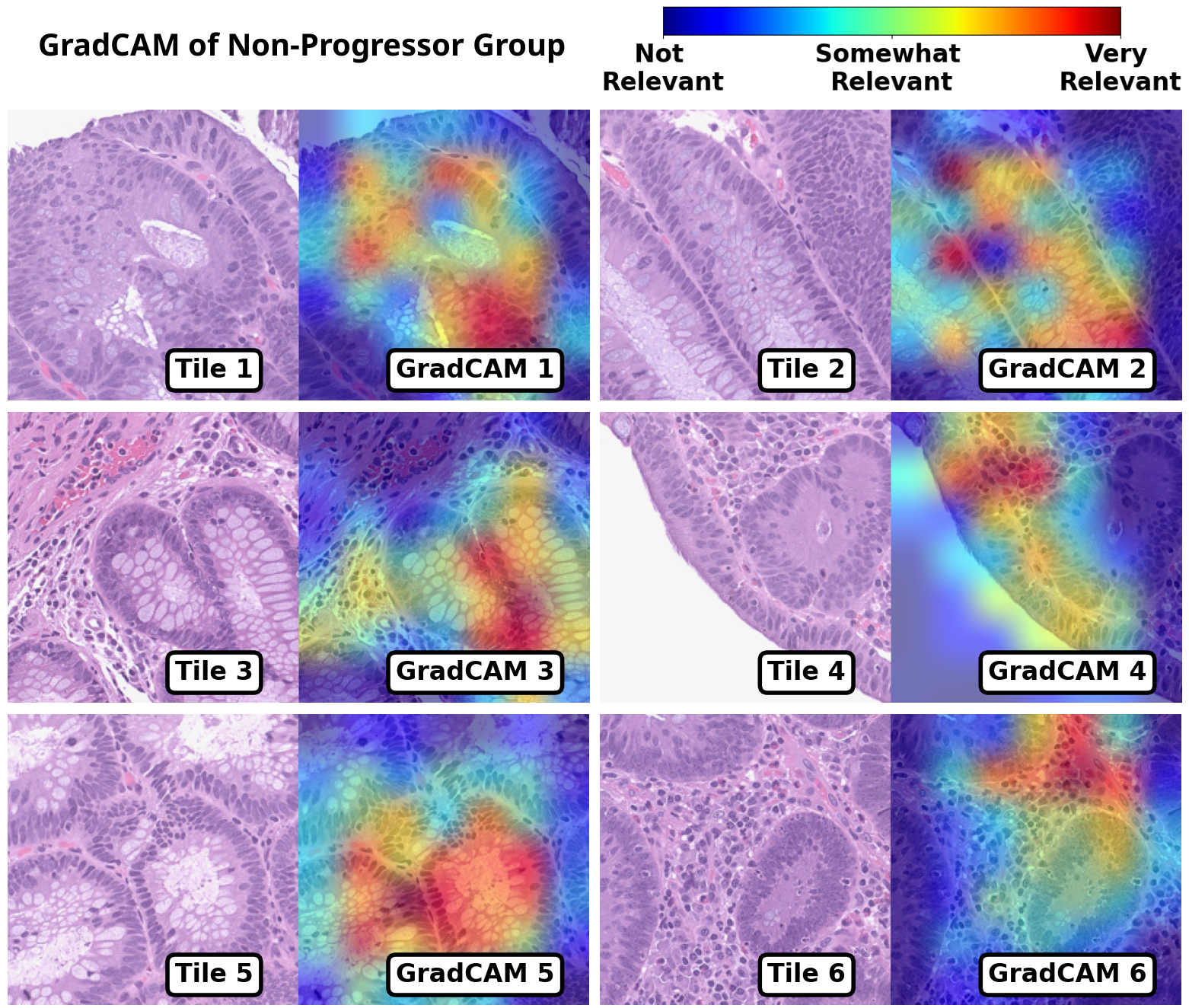}
  \label{fig:4a}
}

\vspace{0.4cm}

\subfloat[Progressor Group – Grad-CAM Heatmap]{%
  \includegraphics[width=0.65\linewidth]{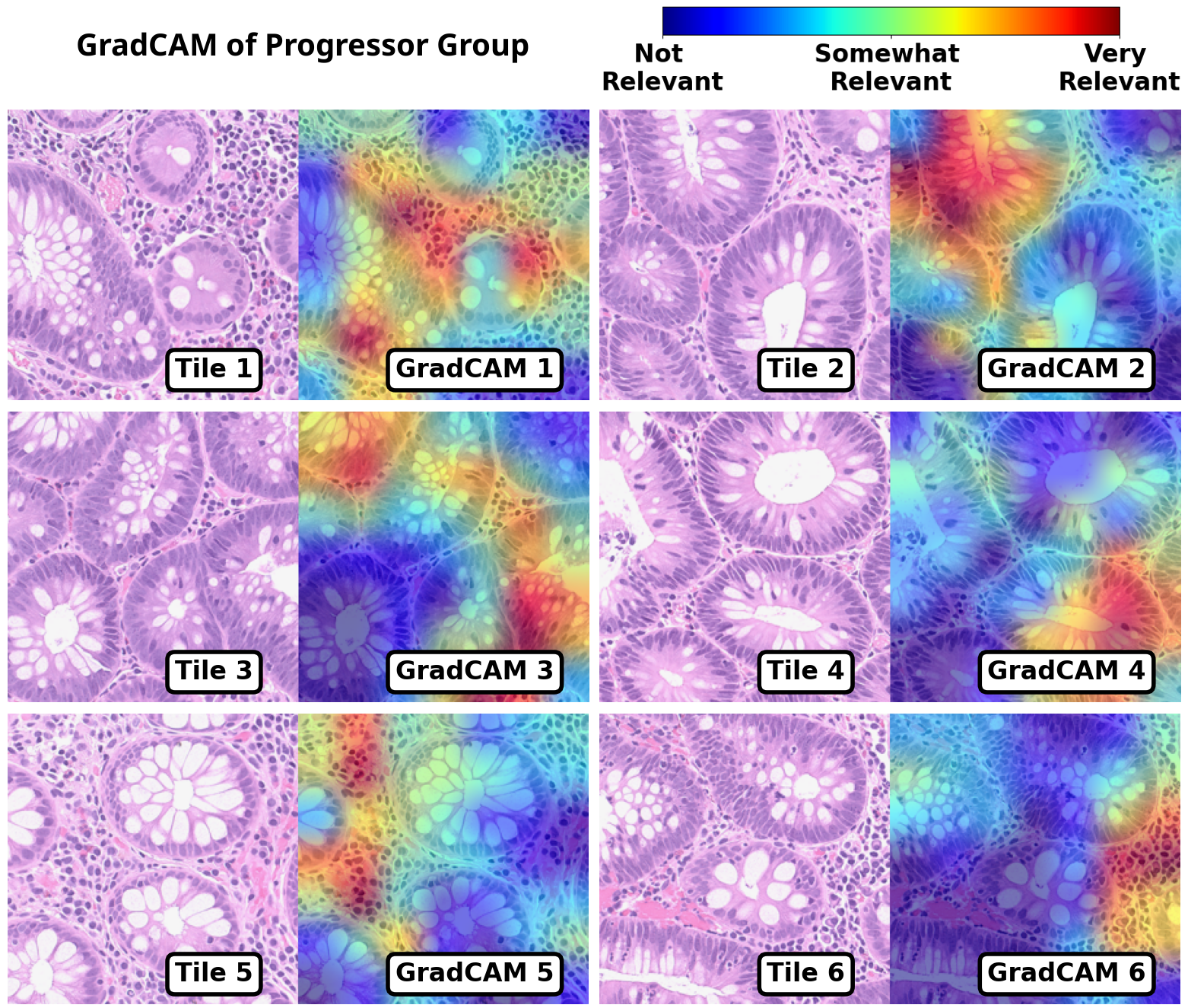}
  \label{fig:4b}
}

\caption{Original Tiles and Grad-CAM Overlays for non-progressor and progressor Groups.  Red indicates the highest relevance for both groups.
(a) Non-progressor group tile-level Grad-CAM heatmap.
(b) Progressor group tile-level Grad-CAM heatmap.}
\label{fig:4}
\end{figure}

To improve interpretability and gain insight into the histologic regions most influential in classification, Gradient-weighted Class Activation Mapping (Grad-CAM) was utilized from the held-out WSIs in both non-progressor and progressor groups \cite{gradcam}. Tiles from one progressor and one non-progressor from the whole WSI evaluation set (unseen during training) were fed through the Grad-CAM algorithm. This serves to visualize the spatial regions within each tile that most strongly influence the model’s predictions. The Grad-CAM algorithm generates a \(7 \times 7\) cell heatmap based on the contribution of the final convolutional layer to the prediction outcome, with red highlighting the regions of highest relevance, offering insights into which histologic features are associated with increased colorectal cancer risk. The yellow-green spectrum range highlights features of intermediate relevance, and the blue features of minimal relevance. 

Figure \ref{fig:4} showcases examples of original tiles paired with their Grad-CAM overlays in which non-progressor (Figure \ref{fig:4a}) and progressor (Figure \ref{fig:4b}) groups are displayed separately. All tiles in Figure \ref{fig:4} were predicted correctly at the tile level. In the non-progressor group, the model primarily focused on areas in which the glandular morphology is preserved: well-formed crypt contours, uniform epithelial lining, and regular nuclear spacing. Furthermore, less-informative regions such as the lamina propria or empty background spaces are highlighted in blue. It is likely that these regions reflect features associated with stable architecture and thus lower potential towards neoplastic progression. In the progressor group, the areas of highest relevance (highlighted red) are localized to regions demonstrating quite some architectural and/or cytological complexity; the red regions highlight crowded, elongated nuclei, irregular glandular outlines, and pseudostratification of the nuclei, being perhaps the main feature highlighted. While these may be considered morphological features of tubular adenomas, these subtle alterations may be indicative of transitions suggestive of early morphologic disorganization and thus may be associated with neoplastic progression. All illustrated tiles were confidently predicted as cancer-associated at the tile level. Together, these heatmaps provide a visual rationale for the model’s decision making and may serve as a tool for hypothesis generation around histological patterns that are not as obvious that may correlate with future cancer development.

\section{Discussion}
The adenoma-carcinoma sequence is perhaps the most widely accepted pathway in the development of colorectal cancer. Yet, standard histopathological examination of low-grade dysplasia in low-risk patients may not reliably distinguish which lesions harbor a higher risk of progression to malignancy. Prior digital pathology studies have demonstrated that deep learning models can extract prognostic information from histology in a variety of contexts \cite{Rad_Huang_Hosseini_Choudhary_Siezen_Akabari_Jamaspishvili_El-Zammar_Patel_Carello_etal._2025}. However, no previous work has assessed whether subtle morphological patterns within low-grade dysplastic lesions detected during screening colonoscopy carry information predictive of long-term progression towards neoplasia; the present study begins to delve into this unexplored area. The null hypothesis stated that no machine-detectable histologic features within low-grade adenomatous polyps are associated with subsequent colorectal cancer development in patients without known predisposing conditions. 

The model demonstrated enhanced performance on more than 40,000 unseen tiles and achieved improved classification across 20 held-out WSIs. These findings suggest that the patients with adenomas that later developed CRC may carry micro-level or even subvisual morphologic cues that discern them from adenomas in patients who do not progress to CRC. 

Grad-CAM interpretability analysis was implemented to further clarify these distinctions. In the progressor group, the regions the model found most relevant often aligned with morphological patterns such as nuclear crowding, elongation, pseudostratification and cues that display architectural complexity, meaning histological features that still fall within the range of low-grade dysplasia but may also hint at early cancerous lesions. In the non-progressor group, the model focused on morphologically preserved gland architecture, such as uniform epithelial formation, evenly spaced nuclei, and precise glandular contours. The automated contrasting activation patterns in the model suggest that the subtle architectural irregularities and a degree of pseudostratification, features that are not traditionally emphasized in routine grading, may carry a predictive value for long-term CRC risk. This pattern of interpretation indicates that early morphologic deviations, while not meeting the thresholds for diagnostic criteria for high-grade dysplasia, may reflect underlying molecular or structural abnormalities. 

Some of the limitations to note are that the dataset was sourced from a single institution and scanned on a single whole slide imaging platform, raising the possibility of institutional bias or scanner-specific bias. This can limit how broadly the findings can be applied. Differences in morphology, staining practices, cutting artifacts, and scanner optics in different laboratories could also affect model performance and limit generalizability. The number of patients that developed CRC was small, even though a large number of tiles were analyzed, reflecting the rarity of this clinical outcome. Future work would benefit from collecting and incorporating datasets from different institutions, validating across multiple scanners, applying stain normalization strategies and conducting studies to better assess robustness and real-world clinical relevance. 

Despite these limitations, the results highlight the potential of deep learning models to detect subtle histologic features in tubular adenomas that may be overlooked by traditional methods and predict cancer development. Automated approaches like this could eventually help patients at higher risk of cancer by providing tailored surveillance and contribute to more precise screening strategies in CRC prevention.

\section{Conclusion}
In this study, a deep learning based EfficientNetV2S model was implemented to detect distinct morphological features in low-grade tubular adenomas, differentiating patients that develop colorectal cancer from those who do not. By effectively modeling a scalable architecture in conjunction with transfer learning and efficient training methodologies, the model positions itself as a tool that can reveal subtle histologic patterns associated with neoplastic progression. This may lay groundwork for future tools aimed at personalized CRC surveillance that conventional methods may overlook. 
\newline

\noindent \textbf{Acknowledgement}

The authors would like to thank the South Bend Medical Foundation for providing access to the whole-slide images used in this study and for their assistance in the preparation and curation of the dataset. 

Declaration of Generative AI and AI-Assisted Technologies in Manuscript Preparation 

During manuscript preparation, the authors used generative AI–based large language model (LLM) tools to assist with drafting and editing text and improving grammar and clarity. These tools were not used for study design, data analysis, or interpretation. All content was reviewed and approved by the authors, who take full responsibility for the published work. 

 \bibliographystyle{elsarticle-num} 
 \bibliography{cas-refs}

@book{Kumar2018Robbins,
  author = {Kumar, Vinay and Abbas, Abul K. and Aster, Jon C.},
  title = {Robbins Basic Pathology},
  publisher = {Elsevier},
  year = {2022},
  date = {12 Dec},
  edition = {11th},
  ISBN = {9780323790185},
}

@Article{info11020125,
    AUTHOR = {Buslaev, Alexander and Iglovikov, Vladimir I. and Khvedchenya, Eugene and Parinov, Alex and Druzhinin, Mikhail and Kalinin, Alexandr A.},
    TITLE = {Albumentations: Fast and Flexible Image Augmentations},
    JOURNAL = {Information},
    VOLUME = {11},
    YEAR = {2020},
    NUMBER = {2},
    ARTICLE-NUMBER = {125},
    URL = {https://www.mdpi.com/2078-2489/11/2/125},
    ISSN = {2078-2489},
    DOI = {10.3390/info11020125}
}

@INPROCEEDINGS{gradcam,
  author={Selvaraju, Ramprasaath R. and Cogswell, Michael and Das, Abhishek and Vedantam, Ramakrishna and Parikh, Devi and Batra, Dhruv},
  booktitle={2017 IEEE International Conference on Computer Vision (ICCV)}, 
  title={Grad-CAM: Visual Explanations from Deep Networks via Gradient-Based Localization}, 
  year={2017},
  volume={},
  number={},
  pages={618-626},
  doi={10.1109/ICCV.2017.74}}

@article{Rad_Huang_Hosseini_Choudhary_Siezen_Akabari_Jamaspishvili_El-Zammar_Patel_Carello_etal._2025, title={Deep learning for digital pathology: A critical overview of methodological framework}, volume={19}, ISSN={21533539}, url={https://linkinghub.elsevier.com/retrieve/pii/S2153353925001002}, DOI={10.1016/j.jpi.2025.100514}, journal={Journal of Pathology Informatics}, author={Rad, Meghdad Sabouri and Huang, Junze (Vincent) and Hosseini, Mohammad Mehdi and Choudhary, Rakesh and Siezen, Harmen and Akabari, Ratilal and Jamaspishvili, Tamara and El-Zammar, Ola and Patel, Palak G and Carello, Saverio J. and Nasr, Michel R. and Rodd, Bardia}, year={2025}, month=nov, pages={100514}, language={en} }

@inproceedings{NIPS,
author = {Krizhevsky, Alex and Sutskever, Ilya and Hinton, Geoffrey E.},
title = {ImageNet classification with deep convolutional neural networks},
year = {2012},
publisher = {Curran Associates Inc.},
address = {Red Hook, NY, USA},
booktitle = {Proceedings of the 26th International Conference on Neural Information Processing Systems - Volume 1},
pages = {1097–1105},
numpages = {9},
location = {Lake Tahoe, Nevada},
series = {NIPS'12}
}

@INPROCEEDINGS{imagenet,
  author={Deng, Jia and Dong, Wei and Socher, Richard and Li, Li-Jia and Kai Li and Li Fei-Fei},
  booktitle={2009 IEEE Conference on Computer Vision and Pattern Recognition}, 
  title={ImageNet: A large-scale hierarchical image database}, 
  year={2009},
  volume={},
  number={},
  pages={248-255},
  doi={10.1109/CVPR.2009.5206848}}

@misc{TensorFlowDevelopers_2025, title={TensorFlow}, rights={Apache License 2.0}, url={https://zenodo.org/doi/10.5281/zenodo.4724125}, DOI={10.5281/ZENODO.4724125}, abstractNote={TensorFlow is an end-to-end open source platform for machine learning. It has a comprehensive, flexible ecosystem of tools, libraries, and community resources that lets researchers push the state-of-the-art in ML and developers easily build and deploy ML-powered applications.}, publisher={Zenodo}, author={TensorFlow Developers}, year={2025}, month=aug }

@article{Tan_Le_2021, title={EfficientNetV2: Smaller Models and Faster Training}, rights={arXiv.org perpetual, non-exclusive license}, url={https://arxiv.org/abs/2104.00298}, DOI={10.48550/ARXIV.2104.00298}, publisher={arXiv}, author={Tan, Mingxing and Le, Quoc V.}, year={2021} }

@article{Bankhead_Loughrey_Fernández_Dombrowski_McArt_Dunne_McQuaid_Gray_Murray_Coleman_etal._2017, title={QuPath: Open source software for digital pathology image analysis}, volume={7}, ISSN={2045-2322}, url={https://www.nature.com/articles/s41598-017-17204-5}, DOI={10.1038/s41598-017-17204-5}, number={1}, journal={Scientific Reports}, author={Bankhead, Peter and Loughrey, Maurice B. and Fernández, José A. and Dombrowski, Yvonne and McArt, Darragh G. and Dunne, Philip D. and McQuaid, Stephen and Gray, Ronan T. and Murray, Liam J. and Coleman, Helen G. and James, Jacqueline A. and Salto-Tellez, Manuel and Hamilton, Peter W.}, year={2017}, month=dec, pages={16878}, language={en} }

@book{Rosai_2011, edition={10}, title={Rosai and Ackerman’s Surgical Pathology, 10e}, volume={1}, publisher={Elsevier}, author={Rosai, Juan}, year={2011}, month=jul, language={en}, ISBN={9780323088046} }

@article{colerectalstats,
author = {Siegel, Rebecca L. and Wagle, Nikita Sandeep and Cercek, Andrea and Smith, Robert A. and Jemal, Ahmedin},
title = {Colorectal cancer statistics, 2023},
journal = {CA: A Cancer Journal for Clinicians},
volume = {73},
number = {3},
pages = {233-254},
doi = {https://doi.org/10.3322/caac.21772},
year = {2023}
}

@article{UPreventive1,
title={Screening for Colorectal Cancer: US Preventive Services Task Force Recommendation Statement}, 
volume={325}, ISSN={0098-7484}, url={https://jamanetwork.com/journals/jama/fullarticle/2779985}, DOI={10.1001/jama.2021.6238}, number={19}, journal={JAMA}, author={US Preventive Services Task Force and Davidson, Karina W. and Barry, Michael J. and Mangione, Carol M. and Cabana, Michael and Caughey, Aaron B. and Davis, Esa M. and Donahue, Katrina E. and Doubeni, Chyke A. and Krist, Alex H. and Kubik, Martha and Li, Li and Ogedegbe, Gbenga and Owens, Douglas K. and Pbert, Lori and Silverstein, Michael and Stevermer, James and Tseng, Chien-Wen and Wong, John B.}, year={2021}, month=may, pages={1965}, language={en} }

@INPROCEEDINGS{ultralight,
  author={Sultana, Aqsa and Abouzahra, Nordin and Rahu, Ahmed and Shula, Brian and Combs, Brandon and Forchetti, Derrick and Aspiras, Theus and Asari, Vijayan K.},
  booktitle={NAECON 2025 - IEEE National Aerospace and Electronics Conference}, 
  title={UltraLight Med-Vision Mamba for Classification of Neoplastic Progression in Tubular Adenomas}, 
  year={2025},
  volume={},
  number={},
  pages={1-6},
  doi={10.1109/NAECON65708.2025.11235447}}

@article{Vennalaganti_Kanakadandi_Goldblum_Mathur_Patil_Offerhaus_Meijer_Vieth_Odze_Shreyas_etal._2017, title={Discordance Among Pathologists in the United States and Europe in Diagnosis of Low-Grade Dysplasia for Patients With Barrett’s Esophagus}, volume={152}, ISSN={00165085}, url={https://linkinghub.elsevier.com/retrieve/pii/S0016508516353173}, DOI={10.1053/j.gastro.2016.10.041}, number={3}, journal={Gastroenterology}, author={Vennalaganti, Prashanth and Kanakadandi, Vijay and Goldblum, John R. and Mathur, Sharad C. and Patil, Deepa T. and Offerhaus, G. Johan and Meijer, Sybren L. and Vieth, Michael and Odze, Robert D. and Shreyas, Saligram and Parasa, Sravanthi and Gupta, Neil and Repici, Alessandro and Bansal, Ajay and Mohammad, Titi and Sharma, Prateek}, year={2017}, month=feb, pages={564-570.e4}, language={en} }

@article{Elmore_Longton_Carney_Geller_Onega_Tosteson_Nelson_Pepe_Allison_Schnitt_etal._2015, title={Diagnostic Concordance Among Pathologists Interpreting Breast Biopsy Specimens}, volume={313}, ISSN={0098-7484}, url={http://jama.jamanetwork.com/article.aspx?doi=10.1001/jama.2015.1405}, DOI={10.1001/jama.2015.1405}, number={11}, journal={JAMA}, author={Elmore, Joann G. and Longton, Gary M. and Carney, Patricia A. and Geller, Berta M. and Onega, Tracy and Tosteson, Anna N. A. and Nelson, Heidi D. and Pepe, Margaret S. and Allison, Kimberly H. and Schnitt, Stuart J. and O’Malley, Frances P. and Weaver, Donald L.}, year={2015}, month=mar, pages={1122}, language={en} }

@misc{korbar2017deeplearningclassificationcolorectalpolyps,
      title={Deep-Learning for Classification of Colorectal Polyps on Whole-Slide Images}, 
      author={Bruno Korbar and Andrea M. Olofson and Allen P. Miraflor and Katherine M. Nicka and Matthew A. Suriawinata and Lorenzo Torresani and Arief A. Suriawinata and Saeed Hassanpour},
      year={2017},
      eprint={1703.01550},
      archivePrefix={arXiv},
      primaryClass={cs.CV},
      url={https://arxiv.org/abs/1703.01550}, 
}

@INPROCEEDINGS{11102194,
  author={Mahajan, Manish and Malik, Kamal and Singh, Amandeep and Sharma, Bharat and Bansal, Ankit and Mehta, Shiva},
  booktitle={2025 International Conference on Networks and Cryptology (NETCRYPT)}, 
  title={Hybrid CNN-LSTM Networks for Enhanced Lung Cancer Detection and Classification from CT Images}, 
  year={2025},
  volume={},
  number={},
  pages={1726-1731},
  doi={10.1109/NETCRYPT65877.2025.11102194}}

@article{Yao_Zhang_Zhou_Liu_2019, title={Parallel Structure Deep Neural Network Using CNN and RNN with an Attention Mechanism for Breast Cancer Histology Image Classification}, volume={11}, ISSN={2072-6694}, url={https://www.mdpi.com/2072-6694/11/12/1901}, DOI={10.3390/cancers11121901}, number={12}, journal={Cancers}, author={Yao, Hongdou and Zhang, Xuejie and Zhou, Xiaobing and Liu, Shengyan}, year={2019}, month=nov, pages={1901}, language={en} }

@article{Kaddes_Ayid_Elshewey_Fouad_2025, title={Breast cancer classification based on hybrid CNN with LSTM model}, volume={15}, ISSN={2045-2322}, url={https://www.nature.com/articles/s41598-025-88459-6}, DOI={10.1038/s41598-025-88459-6}, number={1}, journal={Scientific Reports}, author={Kaddes, Mourad and Ayid, Yasser M. and Elshewey, Ahmed M. and Fouad, Yasser}, year={2025}, month=feb, pages={4409}, language={en} }

@article{Raju_Jayavel_Rajalakshmi_Rajababu_2025, title={CRCFusionAICADx: Integrative CNN-LSTM Approach for Accurate Colorectal Cancer Diagnosis in Colonoscopy Images}, volume={17}, ISSN={1866-9956, 1866-9964}, url={https://link.springer.com/10.1007/s12559-024-10357-2}, DOI={10.1007/s12559-024-10357-2}, number={1}, journal={Cognitive Computation}, author={Raju, Akella S. Narasimha and Jayavel, Kayalvizhi and Rajalakshmi, Thulasi and Rajababu, M.}, year={2025}, month=feb, pages={14}, language={en} }

@article{Asiri_Senan_Halawani_Abunadi_Mashraqi_Alshari_2025, title={Analyzing histopathological images using fused CNN features based on the geometric active contour method for early diagnosis of lung and colon cancer}, volume={16}, ISSN={2730-6011}, url={https://link.springer.com/10.1007/s12672-025-03907-z}, DOI={10.1007/s12672-025-03907-z}, number={1}, journal={Discover Oncology}, author={Asiri, Yousef and Senan, Ebrahim Mohammed and Halawani, Hanan T. and Abunadi, Ibrahim and Mashraqi, Aisha M. and Alshari, Eman A.}, year={2025}, month=nov, pages={2036}, language={en} }

@article{Steimetz_Simsek_Saha_Xia_Gupta_2025, title={Deep learning model for detecting high-grade dysplasia in colorectal adenomas}, volume={17}, ISSN={21533539}, url={https://linkinghub.elsevier.com/retrieve/pii/S2153353925000264}, DOI={10.1016/j.jpi.2025.100441}, journal={Journal of Pathology Informatics}, author={Steimetz, Eric and Simsek, Zeliha Celen and Saha, Asmita and Xia, Rong and Gupta, Raavi}, year={2025}, month=apr, pages={100441}, language={en} }

@article{McCaffrey_Jahangir_Murphy_Burke_Gallagher_Rahman_2024, title={Artificial intelligence in digital histopathology for predicting patient prognosis and treatment efficacy in breast cancer}, volume={24}, ISSN={1473-7159, 1744-8352}, url={https://www.tandfonline.com/doi/full/10.1080/14737159.2024.2346545}, DOI={10.1080/14737159.2024.2346545}, number={5}, journal={Expert Review of Molecular Diagnostics}, author={McCaffrey, Christine and Jahangir, Chowdhury and Murphy, Clodagh and Burke, Caoimbhe and Gallagher, William M. and Rahman, Arman}, year={2024}, month=may, pages={363–377}, language={en} }

@article{MollaHoseyni_Imany_Iranpour_Mehrabani_Seifouri_Rafieipour-Jobaneh_Firuzbakht_Masoudi-Nejad_2025, title={Bridging traditional and deep learning methods in H\&E histological image normalization: a comprehensive review and introducing a novel framework for comparative analyses}, ISSN={20901232}, url={https://linkinghub.elsevier.com/retrieve/pii/S2090123225007532}, DOI={10.1016/j.jare.2025.09.049}, journal={Journal of Advanced Research}, author={Molla Hoseyni, Behnaz Haji and Imany, Sevda and Iranpour, Ahmadreza and Mehrabani, Maryam and Seifouri, Sina and Rafieipour-Jobaneh, Maryam and Firuzbakht, Sina and Masoudi-Nejad, Ali}, year={2025}, month=oct, pages={S2090123225007532}, language={en} }

@article{Elsheikh_Kirkpatrick_Fischer_Herbert_Renshaw_2010, title={Does the time of day or weekday affect screening accuracy?: A pilot correlation study with cytotechnologist workload and abnormal rate detection using the ThinPrep imaging system}, volume={118}, rights={http://onlinelibrary.wiley.com/termsAndConditions#vor}, ISSN={1934-662X, 1934-6638}, url={https://acsjournals.onlinelibrary.wiley.com/doi/10.1002/cncy.20060}, DOI={10.1002/cncy.20060}, number={1}, journal={Cancer Cytopathology}, author={Elsheikh, Tarik M. and Kirkpatrick, Joseph L. and Fischer, Dan and Herbert, Kristi D. and Renshaw, Andrew A.}, year={2010}, month=feb, pages={41–46}, language={en} }

% \end{thebibliography}
\end{document}